\documentclass[10pt,twocolumn,letterpaper]{article}

\usepackage{wacv}              

\usepackage[accsupp]{axessibility}

\usepackage{graphicx}
\usepackage{amsmath}
\usepackage{amssymb}
\usepackage{booktabs}
\usepackage{algcompatible}
\usepackage{float} 

\usepackage[vlined,linesnumbered,ruled]{algorithm2e}
\usepackage[pagebackref,breaklinks,colorlinks]{hyperref}

\usepackage{adjustbox}
\usepackage{multirow}

\usepackage[belowskip=-2pt,aboveskip=1pt]{caption}
\usepackage[capitalize]{cleveref}
\crefname{section}{Sec.}{Secs.}
\Crefname{section}{Section}{Sections}
\Crefname{table}{Table}{Tables}
\crefname{table}{Tab.}{Tabs.}

\def\wacvPaperID{00120} 
\def\confName{WACV}
\def\confYear{2025}

\begin{document}

\title{Label Calibration in Source Free Domain Adaptation}
\author{Shivangi Rai\hspace{0.75em}  Rini Smita Thakur \hspace{0.75em} Kunal Jangid\hspace{0.75em}  Vinod K Kurmi\\
Indian Institute of Science Education and Research Bhopal, India\\
{\tt\small \{{shivangir23,\hspace{0.01em} rinithakur,\hspace{0.01em} kunal24,\hspace{0.01em} vinodkk}\}@iiserb.ac.in }
}
\maketitle

\begin{abstract}
Source-free domain adaptation (SFDA) utilizes a pre-trained source model with unlabeled target data. Self-supervised SFDA techniques generate pseudolabels from the pre-trained source model, but these pseudolabels often contain noise due to domain discrepancies between the source and target domains. Traditional self-supervised SFDA techniques rely on deterministic model predictions using the softmax function, leading to unreliable pseudolabels. In this work, we propose to introduce predictive uncertainty and softmax calibration for pseudolabel refinement using evidential deep learning. The Dirichlet prior is placed over the output of the target network to capture uncertainty using evidence with a single forward pass. Furthermore, softmax calibration solves the translation invariance problem to assist in learning with noisy labels. We incorporate a combination of evidential deep learning loss and information maximization loss with calibrated softmax in both prior and non-prior target knowledge SFDA settings. Extensive experimental analysis shows that our method outperforms other state-of-the-art methods on benchmark datasets. The code is available at \href{https://visdomlab.github.io/EKS/}{https://visdomlab.github.io/EKS/}.

\end{abstract}

\section{Introduction}
\label{sec:intro}
Deep learning has shown tremendous performance improvement on various computer vision tasks based on two assumptions (i) identical and independent distribution of training and test data (ii) availability of an enormous amount of labeled dataset~\cite{liu2022deep}. However, in the real world setting there is always a constraint on labeled data availability, and there exists a domain discrepancy between training (source) and the test (target) dataset. Unsupervised domain adaptation (UDA) tackles this constraint by the inclusion of labeled source data and unlabeled target data, which effectively encounter the performance degradation of out-of-distribution sample. However, source data is not readily available during the adaptation stage due to privacy concerns and copyright issues. Therefore, source free domain adaptation (SFDA)~\cite{Kurmi_2021_WACV} has access to only the source model and unlabeled target data during adaptation stage. The principal methodologies of SFDA involve image style translation, domain- based reconstruction, self-training, and self-attention~\cite{li2024comprehensive}. The self-training generates pseudolabels of the target and adapts the network with re-training. \\
The uncertainty quantification plays a pivotal role in pseudolabeling based SFDA methods. The predictive uncertainty comprises data, model and distributional uncertainty, which aids in selecting the appropriate high-confidence pixels of the unlabeled target data~\cite{deng2023uncertainty}. The pseudolabel refinement self-training SFDA methods are based on drop-out based uncertainty thresholding, ensemble, Bayesian uncertainty matching, etc ~\cite{LEE2023183}, \cite{han2019unsupervised}, \cite{chen2021source}. Bayesian modeling leverages moments approximation by learning posterior over weights whereas ensemble models utilize various independently trained networks for analytical estimation of the class probability distributions. These methods are computationally inefficient with memory and time constraints. Bayesian neural networks (BNN's) also suffer from the problem of intractable posterior inference, selection of weight prior, and sampling requirement at inference. The predictive uncertainty obtained by the UDA methods relies on the point estimate of the output, which fails to capture the miscalibration caused by the distribution shift between the source and target~\cite{xie2023dirichlet}. To overcome these problems, we propose dirichlet-based uncertainty calibration, which captures distributional uncertainty using evidential deep learning.\\
Evidential Deep Learning (EDL) quantifies distributional uncertainty using the Dempster-Shefer theory of evidence and subjective logic viewpoint~\cite{sensoy2018evidential}. In EDL, dirichlet prior is placed over the output predictions instead of network weights. Thus, it transforms the point estimate into the probability distribution over the simplex. Without the requirement for sampling, a grounded representation of both epistemic and aleatoric uncertainty can be learned by training a neural network to output the hyperparameters of the higher-order evidential distribution~\cite{amini2020deep}. EDL computes predictive uncertainty with a single forward pass, unlike BNN's, thereby curtailing the computational requirements. \\
Apart from distributional uncertainty, the other factor which can lead to erroneous results in the SFDA setting is the translational invariance of softmax. The softmax operator for multi-class classification gives the same results for two samples if there exists a relative relationship between their logits. For example, the results of the softmax remain unaltered when the logits are added or subtracted by the particular constant value. It can lead to miscalibration in the SFDA setting with noisy pseudolabels. The problem of translational invariance can be solved with calibrated softmax and requisite evidential learning based loss as the conventional cross-entropy loss poses an issue of gradient shrinking with calibrated softmax  \cite{zong2024dirichlet}.\\
\vspace{-1em}
\begin{itemize}
\itemsep -0.2em 
\item Inclusion of Dirichlet-based uncertainty calibration for pseudolabel refinement in both prior knowledge based SFDA setting \textbf{(EKS)} and on non-prior SFDA network \textbf{(ES)}. In the prior setting, unary and binary bound is available for the unlabeled training target dataset. 
\item  Introduction of a calibrated softmax function at the adaptation stage to reduce the effect of translation invariance. The network is trained with novel EDL loss incorporating the effect of distributional uncertainty and translation invariance.
\item Analysis of the combination of EDL with information maximization (IM). Experiments validate our hypothesis that combining EDL with IM leads to better adaptation to the target domain.  
\item Extensive experiments are performed on benchmark datasets: DomainNet40, Office-Home, Office-31 and Digits datasets such as MNIST, SVHN, and USPS, which show performance enhancement over other state-of-the-art methods.
\end{itemize}

\section{Related Work}
\subsection{Pseudolabeling in SFDA}
Pseudolabeling is one of the important SFDA self-supervised self training method. The psedolabeling approach is divided into three stages: prototype generation, pseudolabel assignment and pseudolabel filtering. DeepCluster~\cite{caron2018deep} is the popular method to generate prototype of unlabeled target using K-means clustering. The pseudolabel assignment is usually based on the distance metric between pixel and prototype. The pseudolabels are inevitably noisy as the source model is used for its generation. The filtering or denoising of pseudolabels involve different uncertainty calibration techniques as given in~\cite{litrico2023guiding}, \cite{lu2023uncertainty}, \cite{chen2021source}, \cite{roy2022uncertainty}, \cite{UPL-SFDA}. DPL~\cite{chen2021source} generates pseudolabels by prototype estimation followed by class and pixel level uncertainty calibration. UPL-SFDA~\cite{UPL-SFDA} quantify uncertainty of pseudolabels with duplication of pre-trained source model head mulitple times with perturbations. The other SFDA setting involves loss reweighting with uncertainty calibrated reliable pseudolabels in self-supervised contrastive network. SHOT~\cite{liang2020we} is the benchmark SFDA work with self-supervised pseudolabeling and information maximization. \\
The recent settings of SFDA explores prior knowledge and active selection of target samples to incorporate more information about unlabeled target. In real world scenario, prior knowledge of target is available in terms of constraints on features, predictions, annotation, class proportion, etc. 
KSHOT~\cite{sun2022prior} further modify the SHOT settings for the more realistic real world adaptation problem by inclusion of prior knowledge by putting constraints on unary and binary bound. POST~\cite{raychaudhuri2023prior} is prior guided self-training SFDA network for human pose estimation. It imposes prediction and feature consistency between the student and teacher for adaptation with human pose prior. Future research directions aim to reduce computational constraints due to the requirement of memory bank and Bayesian sampling. 

\subsection{Evidential Deep Learning}
EDL explicitly models uncertainty on the basis of evidence for various computer vision tasks such as classification, regression, segmentation and object detection~\cite{amini2020deep}, \cite{sensoy2018evidential}, \cite{ulmer2021prior}, \cite{park2023active}. It has shown unprecedented success in showing robustness against out-of-distribution samples and adversarial attacks. EDL leverages higher-order evidential prior over likelihood distribution to effectively capture aleatoric, epistemic and distributional uncertainty. The output of evidential network gives prediction along with evidence. EDL effectively differentiates between different sources of uncertainty as it can extricate (a) confilicting evidence (b) lack of evidence. In evidential classification networks, dirichlet prior is placed over multinomial distribution, whereas in regression networks, normal inverse gamma prior is placed over Gaussian distribution. The training methodology of EDL includes evidential losses in collaboration with the heuristic evidence regularization for uncertainty guidance~\cite{pandey2023learn}. In~\cite{bao2021evidential}, EDL utilizes contrastive debiasing for open set action recognition problem to counter static bias and over-confident predictions. DEED~\cite{ashfaq2023deed} introduce EDL in multilabel classification target setting (multilabel MNIST) by mapping between evidential space and embedding space. The deep evidential fusion network perfoms multi-modal medical image classification~\cite{xu2022deep}.\\
Recently, federated evidential active learning has been introduced with both aleatoric and epistemic uncertainty calibration along with diversity relaxation strategy for image segmentation~\cite{chen2024think}. EDL has also been incorporated in continual learning setup to perform incremental object classification and out-of-distribution detection~\cite{aguilar2023continual}. Although, EDL has been explored for classification, regression, federated learning, continual learning, meta-learning, its scope is not analyzed for SFDA settings.

\section{Problem Statement}
We address SFDA problem doing $K$ class classification using the pre-trained source model ($f_{S}$) trained with labeled source data $\lbrace x_i^s, y_i^s\rbrace _{i=1}^{n_s}$ under the domain $D_s$, unlabeled target domain data $\lbrace x_i^t\rbrace _{i=1}^{n_t}$ under domain $D_{t}$ with  prior knowledge about target domain distribution($P_{K}$). The inclusion of prior knowledge is done using unary bound (UB) and binary relationship (BR) of the unlabeled  target data following the convention of~\cite{sun2022prior}. Here, the objective of\textbf{ Evidential KSHOT(EKS)} is to learn the target model ($f_{T}$) and infer $\lbrace y_i^t\rbrace _{i=1}^{n_t}$ using pre-trained source model and unlabeled target data with prior information and the objective of\textbf{ Evidential SHOT(ES)} is to learn the target model ($f_{T}$) and infer $\lbrace y_i^t\rbrace _{i=1}^{n_t}$ using pre-trained source model$(f_{S})$ and unlabeled target data$(D_{t})$, under self-supervised setting without the use of prior knowledge.
\begin{figure*}
    \centering
    \includegraphics[width=0.75\linewidth]{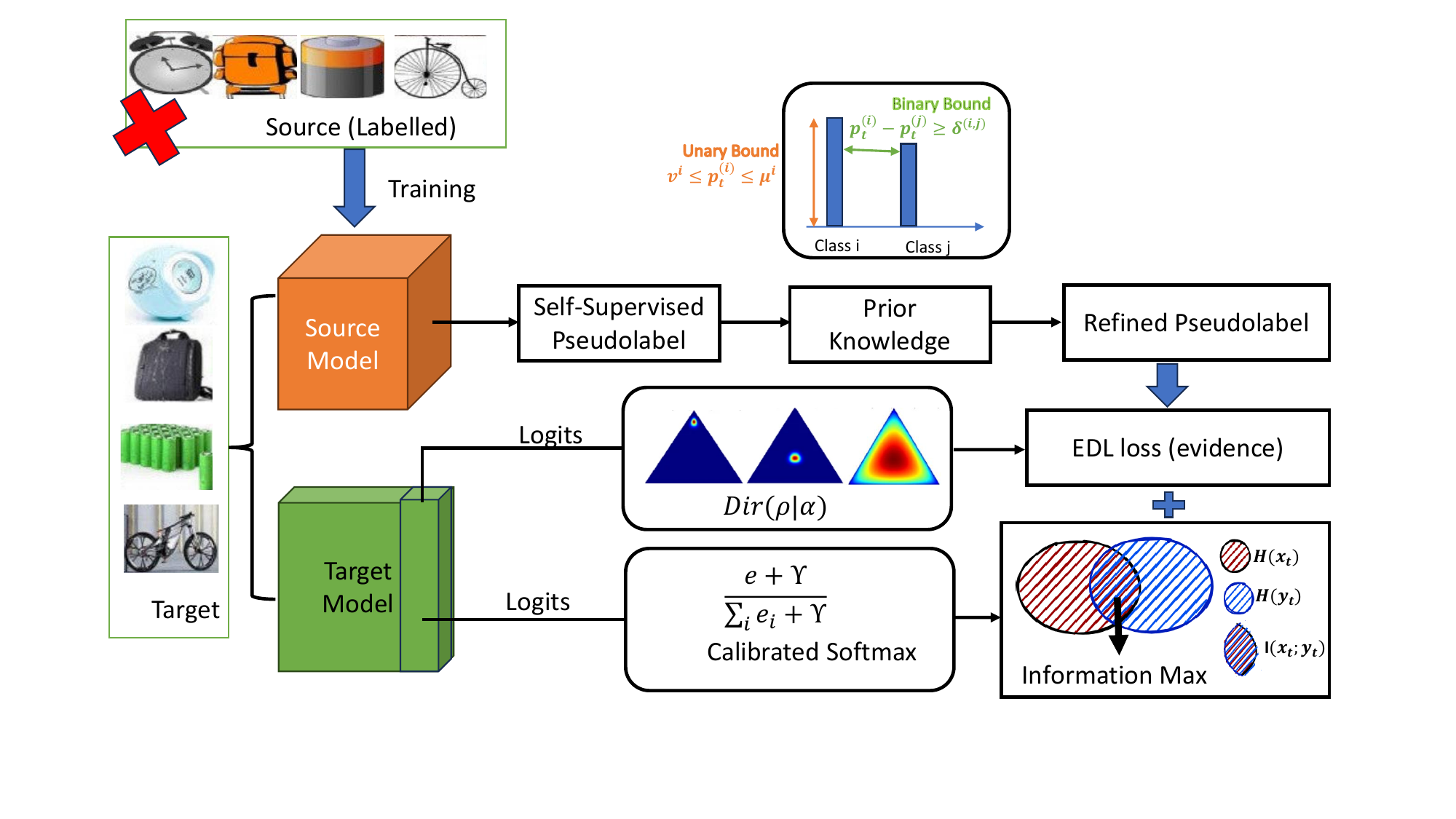}
    \vspace{-2.8em}
    \caption{Overview of the proposed EKS method. The dirichlet distribution and calibrated softmax is placed over logits of the target model followed by adaptation with EDL loss and IM loss in prior knowledge SFDA setting. In the proposed ES method, there is non-availability of prior knowledge, and self-supervised pseudolabels are directly used for EDL loss (best view in color).}
    \label{fig:eks_model}
    \vspace{-1em}
\end{figure*}

\section{Methodology}
\label{sec:method}
The source model ($f_{S}$) is trained using the labeled source data with label smoothing cross entropy objective. The adaptation of the target model relies on the self-training with  pseudolabels generated from the source. We generate the pseudolabels using the prototype estimation and further refine them with prior knowledge as given in KSHOT~\cite{sun2022prior}. The model is further adapted with EDL and calibrated IM as elaborated in Section~\ref{sec:DB} and~\ref{sec:CS}. Figure~\ref{fig:eks_model} illustrates the proposed EKS method which demonstrates the evidential modeling through Dirichlet based calibration and its integration with information maximization.

\subsection{Prior Knowledge}
 The pseudolabels obtained by the pre-trained source model are further refined by prior knowledge constraints of unary bound and binary relationship as given below:
\begin{enumerate}
\item \textbf{Binary Relationship}: It represents the relationship between two target classes.
    \[
    \{(p_t^{(k_i)} - p_t^{(k_{i+1})} \geq 0) \mid i \in \{0, 1, \ldots, K-2\}\}
    \]
\item  \textbf{Unary Bound knowledge creation}: 
    \[
    \{a_k \cdot (1 - \sigma) \leq p_t^{(k)} \leq a_k \cdot (1 + \sigma) \mid k \in \mathcal{K}\}
    \]
\end{enumerate}


where $ \sigma \in$ \{0.0, 0.1, 0.5, 1.0, 2.0\}, $a_{k}$ is the empirical probability of the $k^{th}$ class,   $p_t^{(k)}$ is the class probability of the target distribution , $K$ represents total number of classes, $\sigma$ controls the degree of correctness of prior knowledge estimation.

As we keep increasing the value of $\sigma$ from 0.0 to 2.0, we keep introducing discrepancy in the quality of prior knowledge which leads to significant difference between the distribution of pseudo labels and the ground truth. Due to lack of knowledge under domain shifts, the performance of deterministic models deteriorates as they cannot accurately quantify the uncertainty associated with the new, unseen data distribution. Thus, we incorporate a Dirichlet based prediction calibration based on Evidential Deep Learning (EDL) along with calibrated information maximization for prior knowledge guided SFDA.

\subsection{Dirichlet based uncertainty calibration for knowledge guided SFDA}
\label{sec:DB}
Under domain shifts, the performance of deterministic models deteriorates as they cannot accurately quantify the uncertainty associated with the unseen data distribution of the target domain. Predictive uncertainty is important as it shows how confident a model is in its predictions, especially when there is domain shift. Moreover, self-supervised pseudo-labels require predictive uncertainty calibration for its refinement. 
The Dirichlet-based model predicts class probability by placing a dirichlet distribution over the class probabilities on target model, rather than treating prediction as single point estimate.
The probability density function of $\rho$ for given target sample $x_i^t$ is expressed as:
\begin{equation}
\begin{aligned}
p(\rho | x_i^t, \theta) &= \text{Dir}(\rho | \alpha_i) = \\
&\begin{cases} 
\frac{\Gamma\left(\sum_{k=1}^{K} \alpha_{i_{k}}\right)}{\prod_{k=1}^{K} \Gamma(\alpha_{i_{k}})} \prod_{k=1}^{K} \rho_k^{\alpha_{i_{k}} - 1}, & \text{if } \rho \in \Delta^K \\ 
0, & \text{otherwise}
\end{cases}
\end{aligned}
\label{eq:drich}
\end{equation}
\newline
where, $\alpha_{i}$ is the parameter of the Dirichlet distribution for $x_i^t$, $\Gamma(.)$ is the gamma function, $K$ is total number of classes and 
$\Delta^K = \left\{ \rho_i \bigg| \sum_{k=1}^K \rho_{i_{k}} = 1 \text{ and } \forall j \ 0 \leq \rho_{i_{k}} \leq 1 \right\} $is  a $K$-dimensional unit simplex, where $\rho_i$ is the probability density function for the $i^{th}$ sample and $\rho_{i_{k}}$ denotes probability density function for sample $i$ corresponding to class $k$. EDL modifies the loss function with evidence, which basically deduces the confidence of a particular prediction on the basis of the amount of support collected from data.

The predicted probability of target sample $x_i^t$ for class $k$ can be obtained as: 
\begin{equation}
\begin{aligned}
P(y = k | x_i^t, \theta) &= \int p(y = k | \rho_i)p(\rho_i | x_i^t, \theta) d\rho_i \\
&= \frac{\alpha_{i_{k}}}{\sum_{k=1}^K \alpha_{i_{k}}} \\
&= \frac{exp(o_{i_{k}}) + \lambda}{\sum_{k=1}^K (exp(o_{i_{k}}) + \lambda)}.
\end{aligned}
\end{equation}
here, $\alpha_{i_{k}} = exp(o_{i_{k}}) + \lambda $, 
$o_{{i_{k}}}$ is the observed logit for $i^{th}$ sample corresponding to $k^{th}$ class and $\lambda$ is a constant.

While adapting the source trained model on the target domain, we minimize the negative log of the marginal likelihood $(L_{nll})$ to incorporate the discriminability in the adapted model. Discrimination-based loss $(L_{nll})$ is defined as:
\vspace{-0.5em}
\begin{equation}
    \begin{aligned}
        \mathcal{L}_{\text{nll}} &= -\frac{1}{N} \sum_{i=1}^N \log P(y = k | x_i^t, \theta) \\
        &= \frac{1}{N} \sum_{i=1}^N \sum_{k=1}^K y_{i_{k}} \left[ \log \left( \sum_{j=1}^K \alpha_{i_{j}} \right) - \log \alpha_{i_{k}} \right]
    \end{aligned}
\label{eq:nll}
\end{equation}

In Eq.~\ref{eq:nll}, $y_{i_{k}}$ denotes the refined pseudolabels with prior knowledge. \\
 Divergence-based loss $(L_{kl})$ minimizes the divergence between the predicted Dirichlet distribution and a target Dirichlet distribution. It reduces the impact of the evidence of the false labels for a target sample by reducing it to zero while adapting the model to the target domain.
 \vspace{-0.7em}
\begin{equation}
\begin{aligned}
\mathcal{L}_{\text{kl}} &= \frac{1}{NK} \sum_{i=1}^N D_{\text{KL}} \left( \text{Dir} (\rho_i | \tilde{\alpha}_i) \| \text{Dir} (\rho_i | 1) \right) \\
&= \frac{1}{NK} \sum_{i=1}^N \left[ \log \left( \frac{\Gamma \left( \sum_{j=1}^K \tilde{\alpha}_{i_{j}} \right)}{\Gamma (K) \prod_{j=1}^K \Gamma (\tilde{\alpha}_{i_{j}})} \right) \right. \\
&\quad + \sum_{k=1}^K (\tilde{\alpha}_{i_{k}} - 1) \left( \psi (\tilde{\alpha}_{i_{k}}) - \psi \left( \sum_{j=1}^K \tilde{\alpha}_{i_{j}} \right) \right) \Bigg]
\end{aligned}
\end{equation}

where, $\tilde{\alpha}_i = \mathbf{y}_i + (1 - \mathbf{y}_i) \odot \alpha_i$, is the drichilet parameter for the complementary labels, $\textbf{1}$ is a vector of K ones, $\odot$ is the element-wise multiplication operator  and $\psi(\cdot)$ represents the digamma function.
\newline
Discrimination-based loss ($L_{nll}$) and Divergence-based loss ($L_{kl}$) together constitutes EDL loss $(L_{edl})$.
\begin{equation}
\begin{aligned}
L_{edl}= L_{nll}+ \beta*L_{kl}
\end{aligned}
\end{equation}
where $\beta$ is a hyperparameter, which balances the two losses. 
\subsection{Calibrated Softmax integration in Information Maximization }
\label{sec:CS}
Entropy-based loss ($L_{ent})$  works towards reducing the uncertainty in each prediction by making the predictions confident. 
\vspace{-0.5em}
\begin{equation}
L_{\text{ent}}(f_t; D_t) = -\mathbb{E}_{d_t \in D_t} \left[ \sum_{k=1}^{K} \delta_k(f_t(d_t)) \log \delta_k(f_t(d_t)) \right]
\end{equation}
where, $D_{t}$ represents complete target domain data, $d_{t}$ is the sample from the target domain, $f_{t}$ is the target domain adapted model, $\delta_k$ represents standard softmax function, $K$ is the total number of classes.

Divergence-based loss ($L_{div})$ penalizes the model when there is lesser diversity in predictions.
\vspace{-0.5em}
\begin{equation}
L_{\text{div}}(f_t; D_t) = \sum_{k=1}^{K} \hat{p}_k \log \hat{p}_k = D_{\text{KL}}(\hat{p}, K^{-1} \mathbf{1}_K) - \log K
\end{equation}
where $\hat{p}_k$ is the average class probability of the $k^{th}$ class. $L_{\text{div}}$ estimates the discrepancy between the empirical distributions of target predictions with uniform distribution, thereby encouraging the target network to make diverse predictions by avoiding bias towards a particular class.

But in cases where the source model is trained on a very small set of data or data from a very different distribution than the target domain, $L_{ent}$ might push the model to make overconfident predictions as we go on minimizing the entropy. Standard softmax (Eq.~\ref{softmax}) performs good on common-closed datasets with clean labels but it fails with erroneous results in the noisy label SFDA setting. Thus, we incorporate a calibrated softmax function (Eq.~\ref{csoftmax} )to tackle the problem of overconfident prediction and also to do away with the translation invariance problem of standard softmax function. The standard softmax $({\delta}_{i_{k}})$ and calibrated softmax $(\hat{\delta}_{i_{k}})$ is given in Eq.~\ref{softmax} and \ref{csoftmax}, respectively.
\vspace{-0.5em}

\begin{align}
{\delta}_{i_{k}} &= \frac{e^{o_{i_{k}}}}{\sum_{j=1}^{K} e^{o_{i_{j}}}}, \quad k = 1, 2, \ldots, K. \label{softmax}\\
\hat{\delta}_{i_{k}} &= \frac{e^{o_{i_{k}}} + \gamma}{\sum_{j=1}^{K} (e^{o_{i_{j}}} + \gamma)}, \quad k = 1, 2, \ldots, K.
\label{csoftmax}
\end{align}
In Eq.~\ref{csoftmax}, $\gamma$ is a calibrated softmax constant, which is a hyperparameter.
Calibrated Entropy-based loss $(\hat{L}_{ent})$ make use of calibrated softmax function $(\hat{\delta}_{i_{k}})$ rather than standard softmax function $(\delta_{i_{k}})$. Similarly, we levarage calibrated softmax function into the Divergence-based loss, thereby getting $(\hat{L}_{div})$.
\begin{equation}
\hat{L}_{\text{ent}}(f_t; D_t) = -\mathbb{E}_{d_t \in D_t} \left[ \sum_{k=1}^{K} \hat{\delta_k}(f_t(d_t)) \log \hat{\delta_k}(f_t(d_t)) \right]
\end{equation}
\vspace{-1.5em}
\begin{equation}
\hat{L}_{\text{div}}(f_t; D_t) = \sum_{k=1}^{K} \hat{p}_{c_k} \log \hat{p}_{c_k} = D_{\text{KL}}(\hat{p}_c, K^{-1} \mathbf{1}_K) - \log K
\end{equation}
where, $\hat{p}_{c_k}$ is the average class probability for the $k_{th}$ class after the integration of calibrated softmax Eq. 
 (\ref{csoftmax}).
\newline
Calibrated Entropy-based loss $(\hat{L}_{ent})$ and Calibrated Divergence-based loss $(\hat{L}_{div})$ together constitute Calibrated Information Maximization (IM) Loss $(\hat{L}_{im})$. It's inclusion into the model is shown in Figure~\ref{fig:eks_model}.
\vspace{-0.5em}
\begin{equation}
\hat{L}_{im} = \hat{L}_{ent} + \hat{L}_{div}
\end{equation}
\vspace{-2em}
\subsection{Combining IM with EDL }
When the domain shift is large, EDL might struggle to handle uncertainty due to lack of evidence in the target domain.
Combining EDL loss $(L_{edl})$ and IM loss$(\hat{L}_{im})$ ,as demonstarted in Figure~\ref{fig:eks_model}, can better handle significant domain shift issues. Total training loss $(L_{total})$ is defined as:
\vspace{-1.7em}
\begin{equation}
\begin{aligned}
L_{total} &= w_{1}{L}_{edl} + w_{2}\hat{L}_{im}\\
&=w_{1}(L_{nll} +\beta L_{kl}) + w_{2}(\hat{L}_{ent} + \hat{L}_{div})
\end{aligned}
\label{totalloss}
\end{equation}

where $w_{1}$ and $w_{2}$ are the weights given to EDL loss $(L_{edl})$ and information maximization loss $(\hat{L}_{im})$ respectively, $\beta$ is a hyperparameter to balance $L_{nll}$ and $L_{kl}$. In EKS and ES, the value of $w_{1}$ and $w_{2}$ is set to 0.3 and 1.0 respectively. The pseudo-code of proposed EKS is given in Algorithm 1.

\begin{algorithm}[t]

\SetAlgoLined
\KwIn{Pre-trained source model ($f_{S}$) is available  trained with labeled source data $\lbrace x_i^s, y_i^s\rbrace _{i=1}^{n_s}$, Unlabeled target data  $\lbrace x_i^t\rbrace _{i=1}^{n_t}$, Prior Knowledge $P_{K}$} 
\KwOut{Output: Adapted target model $f_{T}$}
\For{mini batch $N$ of unlabeled samples $\lbrace x_i^t\rbrace _{i=1}^{n_t} \in D_{t}$}
{
Prediction of pseudolabels:  $\hat{y}_{t}$\\
Refinement with prior knowledge:  $y_{ik}$\\
Modify target model with dirichlet prior on logits as per Eq.~\ref{eq:drich}\\
Replace softmax with calibrated softmax as per Eq.~\ref{csoftmax}\\
Train with EDL loss and calibrated IM loss as per Eq.~\ref{totalloss}
}
\caption{Evidential Deep Learning pseudolabel refinement in SFDA}
\label{one}

\end{algorithm}

\begin{table*}
    \caption{ Classification accuracies  on Domainnet40 }
    \label{tab:result_domainnet40}
\begin{adjustbox}{max width=\textwidth} 
    \centering
    \begin{tabular}{ccc|ccccccccccccc}
    \toprule
         Method  & $\kappa$ & $\sigma$ & C$\rightarrow$S & C$\rightarrow$P & C$\rightarrow$R & R$\rightarrow$S & R$\rightarrow$C & R$\rightarrow$P & S$\rightarrow$C & S$\rightarrow$P & S$\rightarrow$R & P$\rightarrow$S & P$\rightarrow$C & P$\rightarrow$R & Avg. \\
         \midrule
         SHOT\cite{liang2020we}& - & - & 75.5 & 74 & 88.4 &  \textbf{72.8} &  \textbf{79.4} & 75.4 & \textbf{80.5} & 70.8 & \textbf{88.3}  & 76.2 & \textbf{77.7} & \textbf{89.8} & 79.1 \\
          
          \textbf{ES}& - & - & \textbf{75.51}&\textbf{77.69} &\textbf{90.16}  &72.13 &79.23 & \textbf{80.03}& 79.50 & \textbf{79.09} & 88.17 &\textbf{76.30} &75.42 & 89.71 &\textbf{ 80.24} \\
          \midrule
       KSHOT\cite{sun2022prior}& UB & 0.0 &  77.0&  76.4& 91.5 &  75.3& \textbf{83.6} & 77.5 & 80.2 & 70.3 & \textbf{89.7} & 76.3 & 82.3 &91.7  & 81.0 \\
      \textbf{EKS } & UB & 0.0 & \textbf{78.69} & \textbf{82.91} & \textbf{91.82} & \textbf{77.30} & 83.54 & \textbf{84.37} & \textbf{81.88} & \textbf{81.44} & 88.77 & \textbf{78.56} & \textbf{83.62} & \textbf{92.05} & \textbf{83.74} \\

       \midrule
       KSHOT\cite{sun2022prior} & BR & - &  76.4& 73.7 & 89.1 & \textbf{74.3} & 82.1 & 76.8 & 79.1 &70.2  &\textbf{88.8}  & 75.9 & 80.6 & \textbf{91.7 }&  79.9\\
         
       
      \textbf{EKS } & BR & - & \textbf{77.34} & \textbf{80.90} & \textbf{91.08 }& 74.16 & \textbf{83.27} & \textbf{83.52} & \textbf{80.28} & \textbf{79.90} & 87.75 & \textbf{76.88} & \textbf{80.96} & 91.53 & \textbf{82.30} \\

          \bottomrule
    \end{tabular}
    \end{adjustbox}
    \vspace{-1em}
\end{table*}

\begin{table*}
\footnotesize
    \caption{ Classification accuracies on Office-Home }
    \label{tab:result_office_home}
\begin{adjustbox}{max width=\textwidth} 
    \centering
    \begin{tabular}{ccc|ccccccccccccc}
    \toprule
         Method  & $\kappa$ & $\sigma$ & A$\rightarrow$C & A$\rightarrow$P & A$\rightarrow$R & C$\rightarrow$A & C$\rightarrow$P & C$\rightarrow$R & P$\rightarrow$A & P$\rightarrow$C & P$\rightarrow$R & R$\rightarrow$A & R$\rightarrow$P & R$\rightarrow$C & Avg. \\
         \midrule
         SHOT \cite{liang2020we}& - & - &  \textbf{57.1} & 78.1 &81.5  & 68.0 &\textbf{78.2}  & 78.1 & 67.4 & \textbf{54.9} & \textbf{82.2} & \textbf{ 73.3}&  \textbf{84.3}& \textbf{58.8} & 71.8 \\
         \textbf{ES}& - & - & 56.81 & \textbf{78.64} & \textbf{82.05} & \textbf{68.18}&77.82 & \textbf{79.07}&\textbf{ 68.11}& 54.01&  82.02& 73.20&84.06 & 58.26&  \textbf{71.85}\\
         \midrule
         
       KSHOT\cite{sun2022prior}& UB & 0.0 & 58.2 & \textbf{80.0 }& 82.9 & \textbf{71.1} &  \textbf{80.3}& 80.7 & 71.3 & 56.8 & 83.2 &  75.5&  \textbf{86.6}&  60.3&  73.9\\
       \textbf{EKS}& UB & 0.0 & \textbf{58.95} & 79.49 & 82.86 & 70.62 & 80.22 & \textbf{81.48} & \textbf{71.80} & \textbf{57.21} & \textbf{83.76} & \textbf{75.94} & 86.30 & \textbf{62.27} & \textbf{74.24} \\

         \midrule
       KSHOT\cite{sun2022prior} & BR & - & 57.4 & 78.8 & 82.9 & 70.7 &  80& 80.5 &  70.8& 55.0 & 82.8 & 74.6 & 86.0 &59.9  & 73.3 \\
       \textbf{EKS }& BR & - & \textbf{58.46} & \textbf{78.99 }& \textbf{83.27} & \textbf{71.41} & \textbf{80.33} & \textbf{80.84} & \textbf{71.10} & \textbf{56.74} & \textbf{83.39} &\textbf{ 75.48 }& \textbf{86.33} & \textbf{60.90 }& \textbf{73.94} \\

          \bottomrule
    \end{tabular}
    \end{adjustbox}
    \vspace{-1em}
\end{table*}

\section{Experiments}
\noindent\textbf{Datasets:} \textbf{\textit{Office-Home}}\cite{venkateswara2017deep} is a challenging benchmark, which comprises four different
domains: Artistic (A), ClipArt (C), Product (P), and Real-World (R). It consists of 65 classes in each domain.
\newline
\textbf{\textit{Domainnet40}}\cite{peng2019moment} is a large UDA benchmark. It comprises four domains Clipart (C), Painting (P), Real (R), Sketch (S), with each domain consisting of 40 classes.
\newline
\textbf{\textit{Office31}}\cite{saenko2010adapting} consists of three domains, Amazon(A), Webcam (W) and DSLR (D), with 31 classes in each domain.
\newline
\textbf{\textit{Digits Dataset}}\cite{hoffman2018cycada}  MNIST (M), SVHN (S) and USPS (U) datasets, are the standard domain adaptation benchmarks for digits classification tasks. The three adaptation tasks are: $\textbf{(1)}$ the source model trained on SVHN (S) dataset is adapted to the target domain MNIST(M), $\textbf{(2)}$ the source model trained on MNIST(M) dataset is adapted to the target domain USPS (U), and $\textbf{(3)}$ the source model trained on USPS (U) is adapted to target MNIST (M).

\begin{table*}[t]
    \caption{ECE and NLL on Domainnet40}
    \label{tab:ece_result_domainnet40}
    \centering
    \resizebox{\linewidth}{!}{%
    \begin{tabular}{c|c|c|cccccccccccc|c}
    \toprule
        Metric & Method & $\kappa$ & C$\rightarrow$S & C$\rightarrow$P & C$\rightarrow$R & R$\rightarrow$S & R$\rightarrow$C & R$\rightarrow$P & S$\rightarrow$C & S$\rightarrow$P & S$\rightarrow$R & P$\rightarrow$S & P$\rightarrow$C & P$\rightarrow$R & Avg. \\
        \midrule
        \multirow{2}{*}{\textbf{ECE}} 
        & KSHOT\cite{sun2022prior} & BR  & 0.153 & 0.116 & 0.055 & 0.149 & 0.065 & 0.092 & 0.104 & 0.141 & 0.085 & 0.148 & 0.083 & 0.055 & 0.104 \\
        & \textbf{EKS} & BR  & \textbf{0.116} & \textbf{0.089} & \textbf{0.045} & \textbf{0.130} & \textbf{0.044} & \textbf{0.071} & \textbf{0.083} & \textbf{0.106} & \textbf{0.063} & \textbf{0.121} & \textbf{0.062} & \textbf{0.044} & \textbf{0.081} \\
        \midrule
        \multirow{2}{*}{\textbf{NLL}} 
        & KSHOT\cite{sun2022prior} & BR  & 1.338 & 1.080 & 0.495 & 1.325 & 0.741 & 0.881 & 1.036 & 1.311 & 0.753 & 1.316 & 0.935 & 0.477 & 0.974 \\
        & \textbf{EKS} & BR  & \textbf{1.194} & \textbf{0.972} & \textbf{0.460} & \textbf{1.249} & \textbf{0.734} & \textbf{0.832} & \textbf{0.989} & \textbf{1.108} & \textbf{0.594} & \textbf{1.214} & \textbf{0.892} & \textbf{0.439} & \textbf{0.890} \\
        \bottomrule
    \end{tabular}
    } 
    \vspace{-1em}
\end{table*}

\noindent\textbf{Implementation Details}
The experiments are performed on three benchmark datasets Domainnet40\cite{peng2019moment}, Office-Home~\cite{venkateswara2017deep}, Office31~\cite{saenko2010adapting} and Digits~\cite{hoffman2018cycada} with the same hyperparameters as in the baseline~\cite{sun2022prior}. ES/EKS report the mean of the three  random seed values for each task to fairly compare our results with baseline. Pretrained ResNet-50~\cite{he2016deep} network is backbone for performing domain wise adaptation on the above benchmark datasets. A Gurobi optimizer~\cite{nau2022comparison} is used for solving the rectification problem using prior knowledge. EKS/ES report the average classification accuracy over all the adaptation tasks of a particular dataset, for all values of $\sigma $ in unary bound (UB), 
where, $\sigma \in \{0.0, 0.1, 0.5, 1.0, 2.0\}$ and binary relationship (BR) where, $\sigma= 1.0 $, we use $\beta = 0.5$, as the balancing factor between $L_{nll}$ and $L_{kl}$. The different values of $\gamma$ are 1, 1.5, and 0.2 for Domainnet40, Office-Home and Office31 respectively for softmax calibration, based on the extensive experimentation. For Digits dataset, we use $\gamma = 1$, for all the experiments.

\section{Results and Analysis}
\subsection{Results}
\label{sec:results}

Tables \ref{tab:result_domainnet40}, \ref{tab:result_office_home} and \ref{tab:result_office31} lists results of\textbf{ ES and EKS} on three benchmarks datasets Domainnet40, Office-Home and Office31. EKS ans ES consistently achieve better classification accuracies on the target domain and also improved adaptation accuracy over KSHOT\cite{sun2022prior} and SHOT\cite{liang2020we}, respectively. It has been observed that the classification accuracy of \textbf{EKS} is significantly higher than \textbf{ES}, thereby highlighting the impact of prior knowledge even with evidential modeling. There is significant improvement on the tasks where domain gap is quite large. EKS/ES show consistent improvement in performance on both types of prior knowledge Unary Bound (UB) and Binary relationship (BR).\newline
\textbf{Results on Domainnet40}: As listed in Table \ref{tab:result_domainnet40}, for 
 \textbf{UB ($\sigma = 0.0$)}, \textbf{EKS} improves the performance by $\textbf{+2.74 \%}$ and for\textbf{ BR}, it improves the performance by $\textbf{+2.4 \%}$ in comparison to baseline\cite{sun2022prior}. On the adaptation task (C$\rightarrow$P), We observe a significant improvement of approximately $\textbf{+6\%}$. \textbf{ES} achieves the improvement in  performance by $\textbf{+1.14 \%}$ in comparison to baseline SHOT\cite{liang2020we}. 
\newline
\textbf{Results on Office-Home}: As listed in Table \ref{tab:result_office_home}, for \textbf{UB($\sigma = 0.0$)}, \textbf{EKS} improves the performance by $\textbf{+0.34 \%}$ and for\textbf{ BR}, it improves the performance by $\textbf{+0.64\%}$ in comparison to baseline\cite{sun2022prior}. However, average class improvement is less, it is showing considerable improvement on certain domains as evident from Table \ref{tab:result_office_home}.\textbf{ES} achieves the improvement in performance by \textbf{$\textbf{+0.05 \%}$} in comparison to baseline\cite{liang2020we}.

\begin{table}
    \caption{ Classification accuracies on OFFICE-31. }
    \label{tab:result_office31}
\begin{adjustbox}{max width=\linewidth}
    \centering
    \begin{tabular}{ccc|ccccccc}
    \toprule
         Method  & $\kappa$ & $\sigma$ & A$\rightarrow$W & A$\rightarrow$D & W$\rightarrow$A & W$\rightarrow$D & D$\rightarrow$A & D$\rightarrow$W & Avg. \\
         \midrule
         SHOT \cite{liang2020we} & - & - & 90.1 & 94.0 & 74.3 & \textbf{99.9} & \textbf{74.7} &\textbf{ 98.4} & 88.6 \\
          \textbf{ES} & - & - & \textbf{91.45} & \textbf{94.11} & \textbf{75.39} & 99.27 & 74.47 & 97.36 & \textbf{88.67} \\
          \midrule
        KSHOT\cite{sun2022prior} & UB & 0.0  & \textbf{98.5} & 97.6 & \textbf{76.2} & 99.8 & \textbf{75.0} & \textbf{99.0} & \textbf{91.0} \\
        \textbf{EKS} & UB & 0.0 & 96.4 & \textbf{98.1} & 75.1 & 99.8 & 74.5 & \textbf{99.0} & 90.5 \\
          \midrule
           KSHOT\cite{sun2022prior} & BR & -  & 97.1 & 96.9 &   76.1 & 99.8 & 74.0 & 98.8 & 90.5 \\
       
       \textbf{ EKS} & BR & - & 96.5 & \textbf{97.1} & 75.9 & 99.8 & \textbf{74.3 }& 98.7 & 90.4 \\
    \bottomrule
    \end{tabular}
\end{adjustbox}
\end{table}

\begin{table}
    \caption{ Classification accuracies on Digits dataset }
    \label{tab:result_digits}
\begin{adjustbox}{max width=\linewidth}
    \centering
    \begin{tabular}{ccc|ccc}
    \toprule
         Method  & $\kappa$ & $\sigma$ & SVHN$\rightarrow$MNIST & MNIST$\rightarrow$USPS &USPS$\rightarrow$MNIST \\
         \midrule
         \multirow{6}{*}{KSHOT\cite{sun2022prior}}
         & UB & 0.0     &  94.79  & 94.45  & 91.26 \\
         & UB & 0.1    &  93.67 & 94.25  & 90.88 \\
         & UB & 0.5    &  93.46 & 94.27  & 89.56 \\
         & UB & 1.0    &  93.46 & 94.22  & 89.49 \\
         & UB & 2.0    &  93.46 & 94.22  & 89.49 \\
         & BR & -      &  94.33 & 94.29  & 91.15 \\
       
          \midrule
          \multirow{6}{*}{\textbf{EKS}}
         & UB & 0.0   &   \textbf{95.22} &\textbf{94.60}   & \textbf{91.78} \\
         & UB & 0.1   &  \textbf{93.82}  &\textbf{94.59}   & \textbf{91.25} \\
         & UB & 0.5   &         93.36    &\textbf{94.40}   & \textbf{89.80} \\
         & UB & 1.0   &         93.36    & \textbf{94.47}  & \textbf{89.77} \\
         & UB & 2.0   &         93.36    & \textbf{94.50}  & \textbf{89.7}7 \\
         & BR & -     & \textbf{ 94.65}  & \textbf{94.57}  & \textbf{91.40} \\

    \bottomrule
    \end{tabular}
\end{adjustbox}
\end{table}

\noindent\textbf{Results on Office31}: As listed in Table \ref{tab:result_office31}, on Office31 dataset, \textbf{EKS} improves the performance of some challenging adaptation task like for A$\rightarrow$D, by \textbf{$\textbf{+0.5 \%}$}. \textbf{ES} achieves the improvement in performance on A$\rightarrow$W by \textbf{$\textbf{+1.35 \%}$}, on A$\rightarrow$D by \textbf{$\textbf{+0.11 \%}$} and on W$\rightarrow$A by \textbf{$\textbf{+1.09 \%}$}.

\noindent\textbf{Results on Digits Dataset}: As listed in Table \ref{tab:result_digits}, on Digits dataset, \textbf{EKS} improves the performance of some challenging adaptation task like for SVHN$\rightarrow$MNIST, on various values of $\sigma$, such as on UB(0.0) by \textbf{$\textbf{+0.43 \%}$}, and for\textbf{ BR}, it improves the performance by $\textbf{+0.32\%}$. While adapting the model trained on USPS(U) to the target domain(M), consistent improvement in performance has been achieved through EKS over all values of $\sigma$.

\noindent\textbf{Result of E-TransDA on Office31}: We leverage our methodology into other SFDA technique (TransDA) and named it as E-TransDA. The result for the same is reported in the Table \ref{tab:result_TransDA_office31} on Office-31 dataset. We observe a consistent improvement in accuracy across all domains and in average adaptation accuracy as well, thus validating the generalizability of our approach.
\begin{table}[ht]
    \caption{ Classification accuracies on Office31 (E-TransDA)}
    \label{tab:result_TransDA_office31}
    \centering
    \resizebox{\linewidth}{!}{%
        \begin{tabular}{c|cccccc|c}
        \toprule
        Method & A$\rightarrow$D & A$\rightarrow$W & D$\rightarrow$A & D$\rightarrow$W & W$\rightarrow$A & W$\rightarrow$D & Avg. \\
        \midrule
        TransDA \cite{yang2021transformer} & 97.20 & 95.00 & 73.70 & 99.30 & 79.30 & 99.60 & 90.68 \\
        \textbf{E-TransDA (Ours)} & \textbf{97.59} & \textbf{96.48} & \textbf{74.51} & 98.87 & \textbf{80.09} & \textbf{100.00} & \textbf{91.26} \\
        \bottomrule
        \end{tabular}
    }
    \vspace{-1em}
\end{table}

%





\subsection{Analysis}
\subsubsection{Calibration Analysis}
An empirical analysis that quantifies how well the pre-
dicted confidence levels correspond to actual outcomes
across the entire model, specifically, Expected Calibration
Error (ECE) and Negative log likelihood (NLL) for EKS
is evaluated.Table \ref{tab:ece_result_domainnet40} shows the results of the ECE and NLL on Do-
mainnet40 dataset.

\subsubsection{\textbf{Effect of combining IM with EDL}}

The analysis of the effectiveness of combining information maximization with EDL is done using the Office-Home dataset. Table \ref{tab:ablation_im_loss_result_with_edl} validate the result of the proposed framework. A comparison of classification accuracies is presented between the framework that combines EDL with IM $(L_{total})$ and the framework that solely uses EDL loss $(L_{edl})$, excluding IM loss from the training process.

EDL enhances the reliability of pseudo-labels by making confident predictions. In cases where there is a lack of evidence, EDL might lead the model to make overconfident predictions on noisy  pseudo-labels, as it pushes the model to minimize entropy, which hampers the performance of the adapted model. IM complements EDL by its $L_{div}$ component as it encourages diversified predictions among classes ensuring that the model does not become overconfident for a few classes. There is significant improvement in classification accuracies for various values of $\sigma$ when combining EDL with IM , i.e, improvement of $\textbf{+1.94 \%}$ on UB(0.0), $\textbf{+1.91 \%}$ on UB(0.1) etc. and for BR, the improvement in performance by $\textbf{+2.32 \%}$. Thus, combination of EDL with IM achieve maximum increment in accuracy.  Table\ref{tab:ece_result_domainnet40} shows the results of the ECE and NLL  on Domainnet40.

\begin{table}
\centering
\footnotesize
    \caption{ Ablation study of IM loss on Office-Home dataset}
    \label{tab:ablation_im_loss_result_with_edl}
\begin{adjustbox}{max width=\linewidth}
    \centering
    \begin{tabular}{cc|cc}
    \toprule
          $\kappa$ & $\sigma$ & EKS with IM & EKS w/o IM \\
         \midrule
             UB & 0.0  & \textbf{73.93}  & 71.99 \\
             UB & 0.1  & \textbf{73.71}  & 71.80 \\
             UB & 0.5  & \textbf{72.76} & 70.64 \\
             UB & 1.0  & \textbf{72.19} & 70.00 \\
             UB & 2.0  & \textbf{71.92} & 69.71 \\
             BR &  -   & \textbf{73.62}  & 71.30 \\

    \bottomrule
    \end{tabular}
\end{adjustbox}
\end{table}

\begin{table}
\centering
\footnotesize
    \caption{ Ablation study of calibrated softmax based IM loss on Office Home }
    \label{tab:ablation_calibrated_im_loss_result_office31}
\begin{adjustbox}{max width=\linewidth}
    \centering
    \begin{tabular}{ccc|cc}
    \toprule
         $\kappa$ & $\sigma$ & EKS(Standard Softmax) & EKS (calibrated softmax)  \\
         \midrule
          UB & 0.0  & 73.93 & \textbf{74.24} \\
          UB & 0.1  & 73.71 & \textbf{73.98} \\
          UB & 0.5  & 72.76 & 72.72 \\
          UB & 1.0  & 72.19 & 72.01 \\
          UB & 2.0  & 71.92 & 71.66 \\
          BR & -    & 73.62 & \textbf{73.94} \\
    \bottomrule
    \end{tabular}
\end{adjustbox}
\vspace{-1.5em}
\end{table}

\begin{table*}
\smaller
    \caption{ Ablation study of classification accuracies on Office-Home where $\sigma \in \{0.1,0.5,1.0,2.0\} $}
    \label{tab:ablation_result_officehome}
\begin{adjustbox}{max width=\textwidth} 
    \centering
    \begin{tabular}{ccc|ccccccccccccc}
    \toprule
         Method  & $\kappa$ & $\sigma$ & A$\rightarrow$C & A$\rightarrow$P & A$\rightarrow$R & C$\rightarrow$A & C$\rightarrow$P & C$\rightarrow$R & P$\rightarrow$A & P$\rightarrow$C & P$\rightarrow$R & R$\rightarrow$A & R$\rightarrow$P & R$\rightarrow$C & Avg. \\
         \midrule
         \multirow{4}{*}{\rotatebox[origin=c]{90}{KSHOT \cite{sun2022prior}}} 
         & UB & 0.1 &  58.1& \textbf{79.2} & \textbf{83.2} & 70.4 & \textbf{80.0} & 80.7 & 71.4 &  56.5&  83.0&  \textbf{75.6}&\textbf{ 86.0} &  60.8&  73.7\\
       & UB & 0.5 &  57.4& 79.1 & 82.1 & 69.4 & 78.1 & 79.5 & 69.3 &55.2  & 81.8 &  74.0& 85.1 &  60.2& 72.6 \\
       & UB & 1.0 & 57 & 79 & 82.1 &68.6  & 77.8 & 79.3 & 68.4 & 55.1 & 81.7 &  73.5&  84.8&59.3  & 72.2 \\
       & UB & 2.0 &  56.4& 78.7 & 82.1 & 68.3 &  77.8& 79.3 & 67.9 & 54.2 & 81.7 & 73.3 & 84.8 &58.7  & 71.9 \\
       \midrule
       \multirow{4}{*}{\rotatebox[origin=c]{90}{\textbf{EKS}}}
       & UB & 0.1 & \textbf{58.93} & 78.48 & 83.17 & \textbf{70.59} & 79.42 & \textbf{81.29} & \textbf{71.58} & \textbf{57.74} & \textbf{83.44} & 75.59 & 85.43 & \textbf{62.13} & \textbf{73.98} \\
        & UB & 0.5 & \textbf{58.10 }& 78.73 & 81.98 & 68.83 & \textbf{78.31} & 79.34 & 69.26 & \textbf{56.26} & 81.42 & \textbf{74.38} & 84.78 & \textbf{61.29} & \textbf{72.72} \\
        & UB & 1.0 & 57.00 & 78.47 & 82.04 & 68.29 & \textbf{78.12} & 78.42 & \textbf{68.43} & 54.52 & 80.84 & \textbf{74.10} & 84.04 & \textbf{59.87} & 72.01 \\
        & UB & 2.0 & \textbf{56.43} & 78.06 & 82.04 & 67.89 & \textbf{77.95} & 78.44 & \textbf{67.92} & 53.48 & 80.84 & \textbf{73.93} & 84.07 & \textbf{58.83 }& 71.66 \\
        \bottomrule
    \end{tabular}
    \end{adjustbox}
    
\end{table*}

\begin{table*}[t]
\smaller
    \caption{ Ablation study of classification accuracies on Domainnet40 where $ \sigma \in$ \{0.0, 0.1, 0.5, 1.0, 2.0\} }
    \label{tab:ablation_result_domainnet40}
\begin{adjustbox}{max width=\textwidth} 
    \centering
    \begin{tabular}{ccc|ccccccccccccc}
    \toprule
         Method  & $\kappa$ & $\sigma$ & C$\rightarrow$S & C$\rightarrow$P & C$\rightarrow$R & R$\rightarrow$S & R$\rightarrow$C & R$\rightarrow$P & S$\rightarrow$C & S$\rightarrow$P & S$\rightarrow$R & P$\rightarrow$S & P$\rightarrow$C & P$\rightarrow$R & Avg. \\
         \midrule
         \multirow{4}{*}{\rotatebox[origin=c]{90}{KSHOT \cite{sun2022prior}}} 
         & UB & 0.1 & 76.9 & 76.8 & 89.5 &  75.2& 82.2 & 77.6 & 79.4 &  70.1& \textbf{88.5} &  76.9& 81.7 & 91.2 & 80.5 \\
       & UB & 0.5 & 75.6 & 75.4 & 88.8 & 73.5 &  80.3& 77.2 & 78.9 &  70.4&88.3  & 76.6 & 78.4 & 89.0 & 79.4 \\
       & UB & 1.0 & 75.3 & 73.7 & 88.8 & 72.9 & 79.9 & 76.8 & 80.0 & 69.9 & 88 &  76.4&  77.6& 88.6 & 79.0 \\
       & UB & 2.0 & 75.5& 75.4 &  88.8& 73.1 & 79.2 & 76.3 & 80.1 &  71.1&87.9  & 76.2 &  77.8& 88.6 & 79.2 \\
       \midrule
         \multirow{4}{*}{\rotatebox[origin=c]{90}{\textbf{EKS }}} 
       & UB & 0.1 & \textbf{78.88} & \textbf{82.49} & \textbf{91.34} & \textbf{76.78} & \textbf{83.69} & \textbf{83.97} & \textbf{81.06} & \textbf{81.42} & 88.03 & \textbf{78.17} & \textbf{82.36} & \textbf{91.65} & \textbf{83.32} \\
       & UB & 0.5 & \textbf{76.31} & \textbf{79.55} & \textbf{90.62} & \textbf{73.84} &\textbf{ 81.87} & \textbf{83.19} & \textbf{80.03} & \textbf{80.84 }& 87.81 & \textbf{77.20} & 77.84 & \textbf{90.66 }& \textbf{81.65} \\
       & UB & 1.0 & \textbf{75.60} & \textbf{79.04} & \textbf{90.49 }& \textbf{73.23} & 79.60 & \textbf{82.86} & 79.72 & \textbf{80.44} & 87.46 & \textbf{76.73} & 75.60 & \textbf{90.26 }& \textbf{80.92} \\
       & UB & 2.0 & 75.28 & \textbf{78.79 }& \textbf{90.47} & 73.02 & 79.19 & \textbf{82.39} & 79.56 & \textbf{79.55} & 87.44 & \textbf{76.35 }& 75.43 & \textbf{90.27} & \textbf{80.64} \\
          \bottomrule
    \end{tabular}
    \end{adjustbox}
    \vspace{-1.2 em}
\end{table*}

\begin{table}
\footnotesize
    \caption{ Ablation study of classification accuracies on Office-31 where, $ \sigma \in$ \{0.0, 0.1, 0.5, 1.0, 2.0\} }
    \label{tab:ablation_result_office31}
\begin{adjustbox}{max width=\linewidth}
    \centering
    \begin{tabular}{ccc|ccccccc}
    \toprule
         Method  & $\kappa$ & $\sigma$ & A$\rightarrow$W & A$\rightarrow$D & W$\rightarrow$A & W$\rightarrow$D & D$\rightarrow$A & D$\rightarrow$W & Avg. \\
         \midrule
         \multirow{4}{*}{\rotatebox[origin=c]{90}{KSHOT \cite{sun2022prior}}} 
         & UB & 0.1  & \textbf{97.2} & \textbf{96.7} & \textbf{76.5} & 99.8 & 75.5 & \textbf{98.7} & \textbf{90.7} \\
         & UB & 0.5  & 92.8 & 93.9 & 75.2 & 99.7 & 75.7 & 97.7 & 89.2 \\
         & UB & 1.0  & 92.4 & 93.7 & 75.5 & 99.7 & \textbf{75.5} & 97.7 & 89.1 \\
         & UB & 2.0  & 92.4 & 93.7 & 75.2 & 99.7 & 75.0 & 97.7 & 89.0 \\

          \midrule
         \multirow{4}{*}{\rotatebox[origin=c]{90}{\textbf{EKS}}} 
         & UB & 0.1  & 96.4 & 96.6 & 75.8 & \textbf{99.8} & 75.3 & 98.3 & 90.4 \\
         & UB & 0.5 & \textbf{93.2 }& \textbf{94.5} & 75.0 & 99.5 & 75.3 & 97.4 & 89.2 \\
         & UB & 1.0 & 91.4 & \textbf{94.4} & 74.8 & 99.5 & 74.6 & 97.4 & 88.7 \\
         & UB & 2.0  & 91.3 & \textbf{94.4} & 74.8 & 99.5 & 74.7 & 97.4 & 88.7 \\
    \bottomrule
    \end{tabular}
\end{adjustbox}
\vspace{-1em}
\end{table}

\subsubsection{\textbf{Integrating calibrated softmax function with IM}}
Table \ref{tab:ablation_calibrated_im_loss_result_office31} validates the effectiveness of integrating the calibrated softmax function Eq.$(\ref{csoftmax})$ instead of the standard softmax function Eq.$(\ref{softmax})$. It reports comparison of classification accuracies on Office-Home dataset between standard softmax integration with IM and calibrated softmax integration with IM. Calibrated softmax function mitigates the issue of translation invariance faced by the standard softmax function and predicts the probabilities by keeping in consideration the magnitude of the logits but the standard softmax only predicts the probabilities based on the relative relationships of the logits and does not consider the effect of the magnitude of the logits in the predictions. As reported in Table \ref{tab:ablation_calibrated_im_loss_result_office31}, there is consistent improvement in classification accuracies such as by $\textbf{+0.31 \%}$ for UB(0.0) and by $\textbf{+0.32 \%}$ for BR on Office-Home dataset with integration of the calibrated softmax function with the information maximization. 

\vspace{-1em}

\subsubsection{\textbf{Effect of proposed EKS on various values of ($\sigma$)}}
As the value of $\sigma$ grows from 0.1 to 2.0,  the certainty of prior knowledge gets diluted more and more. Thus, with less knowledge, it becomes a challenge to improve the adaptation. Our proposed framework also performs better in these cases. The maximum increment in accuracy is at UB 0.1. The ablation study on classification accuracy at different value of bounds $(\sigma)$ for Domainnet40, Office-Home and Office31 are given in  Table \ref{tab:ablation_result_officehome} \ref{tab:ablation_result_domainnet40}, \ref{tab:ablation_result_office31} respectively.  
\vspace{-1 em}

\begin{figure}
  \centering
  \begin{subfigure}{0.49\linewidth}
    \includegraphics[width=\linewidth]{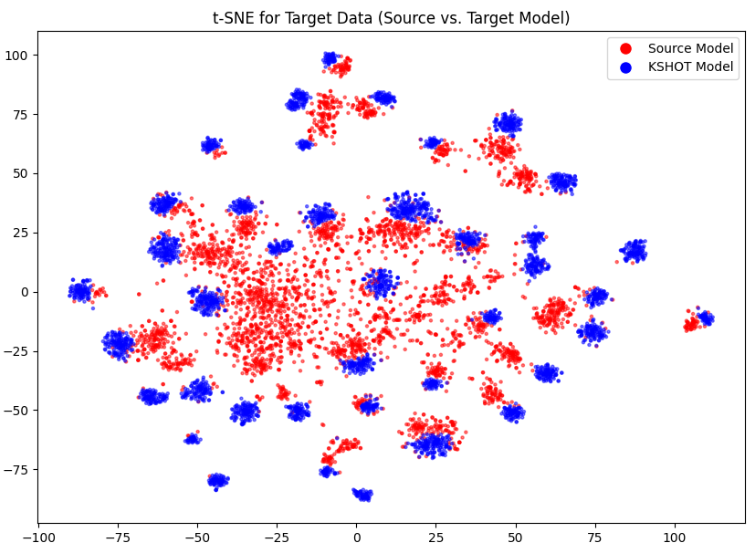}
    \caption{P$\rightarrow$C (KSHOT)}
    \label{fig2(a):}
  \end{subfigure}
  \hfill
  \begin{subfigure}{0.49\linewidth}
    \includegraphics[width=\linewidth]{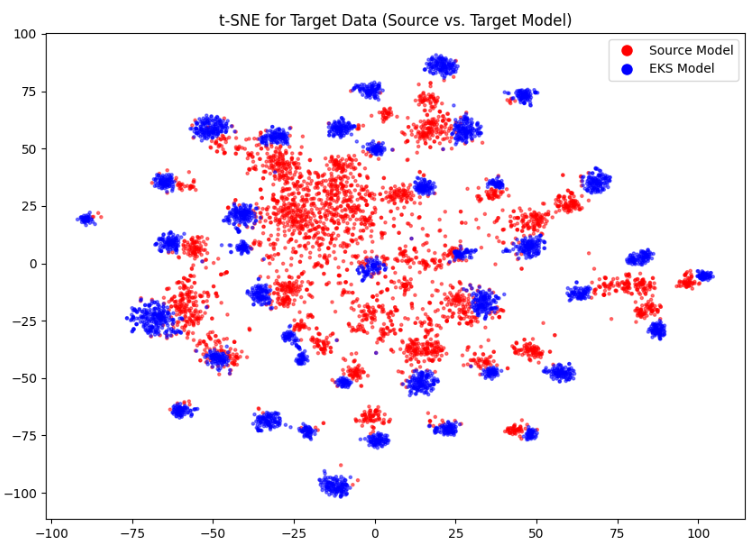}
    \caption{P$\rightarrow$C (EKS)}  
    \label{fig2(b):}
  \end{subfigure}

  \begin{subfigure}{0.49\linewidth}
    \includegraphics[width=\linewidth]{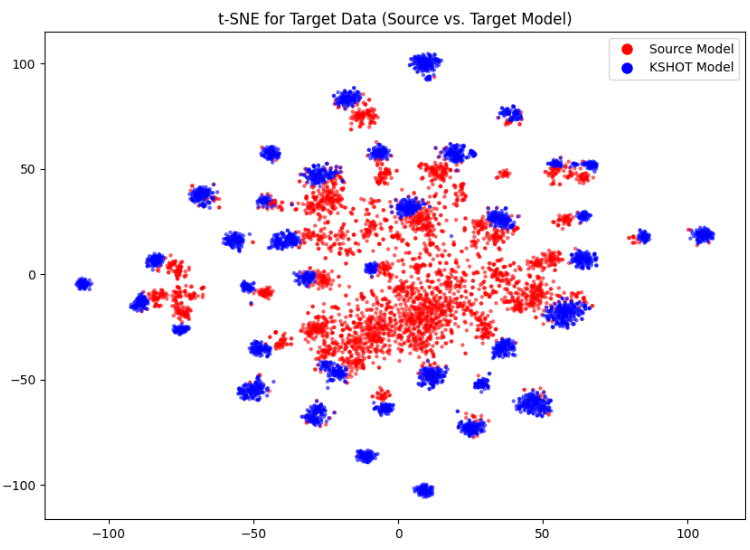}
    \caption{S$\rightarrow$C (KSHOT)}
    \label{fig(3a):}
  \end{subfigure}
  \hfill
  \begin{subfigure}{0.49\linewidth}
    \includegraphics[width=\linewidth]{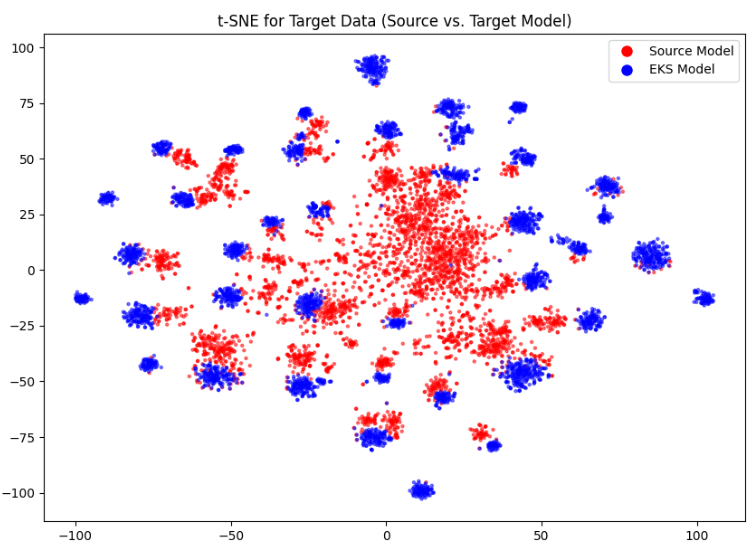}
    \caption{S$\rightarrow$C (EKS)}  
    \label{fig3(b):}
  \end{subfigure}
  \
  \caption{t-SNE plots for P$\rightarrow$C and S$\rightarrow$C on Domainnet40.}
  \label{fig3:}

  \vspace{-1.2 em}
\end{figure}

\subsubsection{Feature and Calibration Visualization}
\vspace{-0.5 em}

\begin{figure}[ht]
  \centering

  \begin{subfigure}{0.49\linewidth}
    \includegraphics[width=\linewidth]{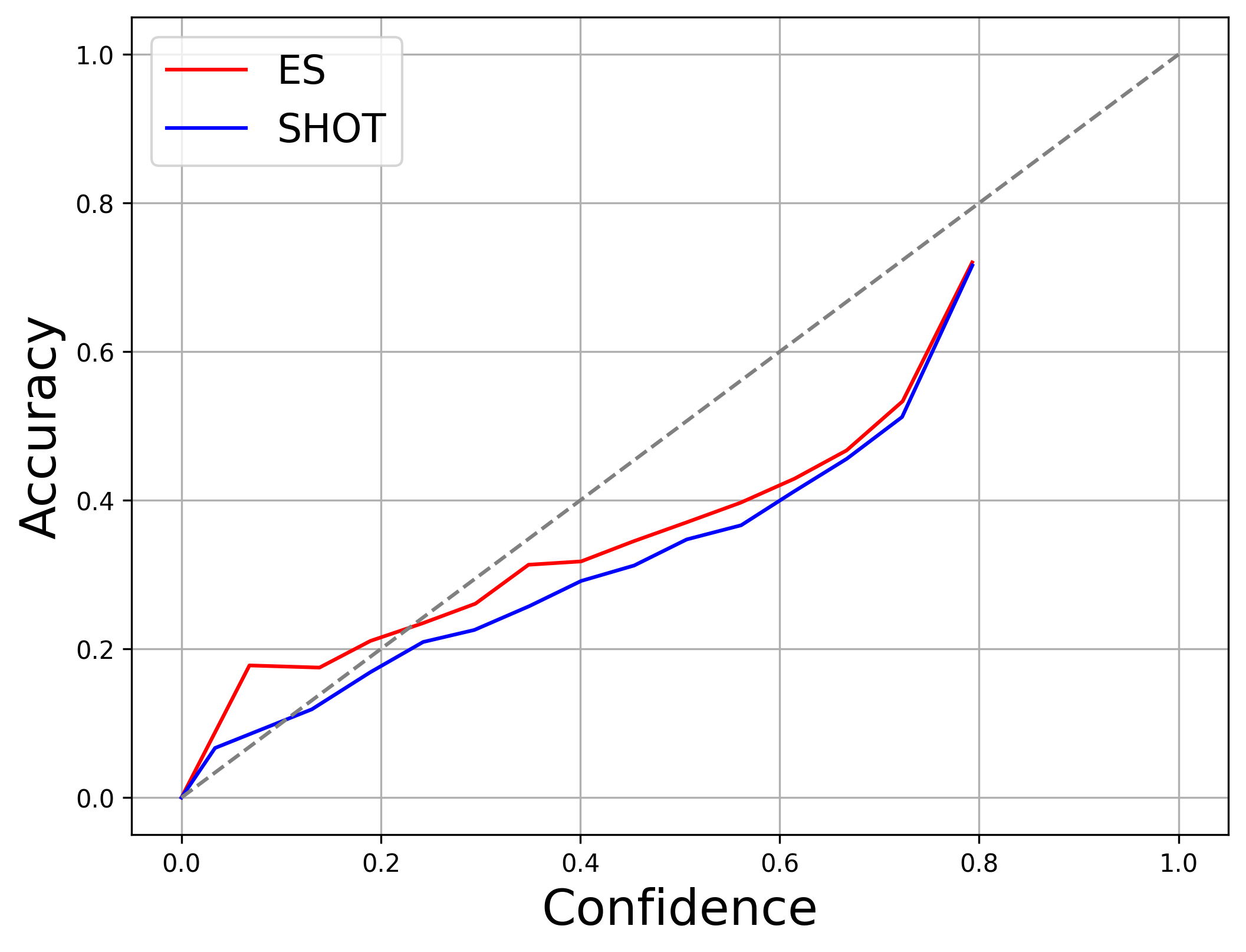}
    \caption{ES vs SHOT}
    \label{fig:ECE_ES_SHOT}
  \end{subfigure}
    \hfill
  \begin{subfigure}{0.49\linewidth}
    \includegraphics[width=\linewidth]{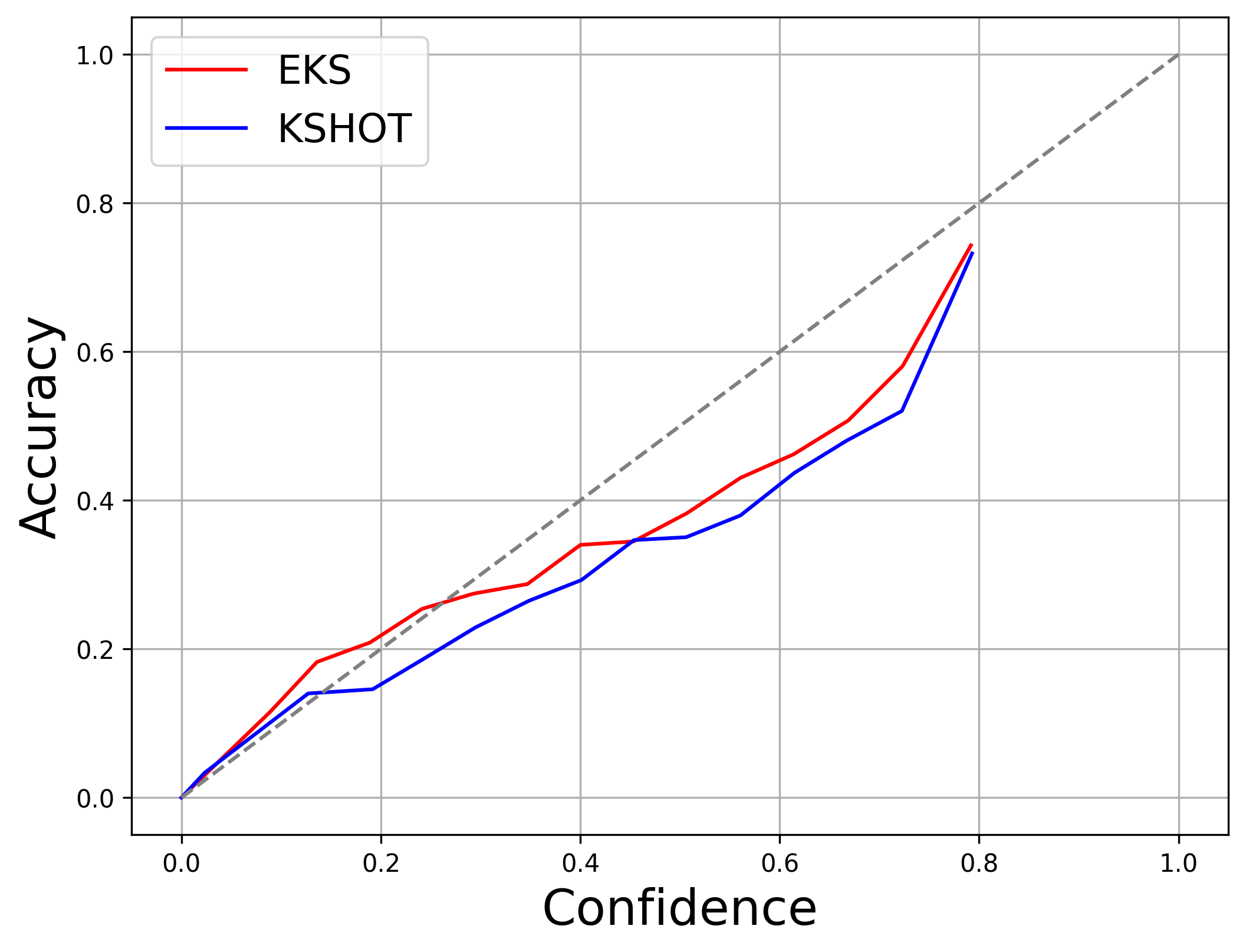}
    \caption{EKS vs KSHOT}
    \label{fig:ECE_EKS_KSHOT}
  \end{subfigure}
    \vspace{0.35em}
    \caption{Expected Calibration Error (ECE) curve  on the Domainnet40 dataset.}
    \vspace{-1em}
  \label{fif_ECE_overall}
\end{figure}

Figure~\ref{fig2(a):} and~\ref{fig2(b):} show the t-SNE visualization of features learned by the model when trained with KSHOT and EKS respectively, on  task (P$\rightarrow$C) , while Figure~\ref{fig(3a):} and~\ref{fig3(b):} show the t-SNE visualization of features learned by the model when trained with KSHOT and EKS  respectively on adaptation task (S$\rightarrow$C). It is observed that adaptation of features achieved through EKS is better than baseline KSHOT\cite{sun2022prior}. 
\noindent {\bf Calibration curve:} Figure \ref{fif_ECE_overall} shows the ECE curve for visualizing calibration, where the dashed line $(x = y)$ represents ideal calibration. A curve closer to this line indicates better calibration. In Figure \ref{fig:ECE_EKS_KSHOT}, the EKS curve is closer to the dashed line compared to KSHOT, while Figure \ref{fig:ECE_ES_SHOT} shows the ES curve is closer to the dashed line than SHOT, thus, validating our approach.

\vspace{-1.0em}

\section{Conclusion}
In this paper, SFDA has been addressed in two different settings: EKS (with prior knowledge) and ES (w/o prior knowledge). An evidence-based model training is implemented, assigning an input sample to the class with the strongest evidence. The effect of using a calibrated softmax function for predicting probability and its integration into information maximization has been explored. Extensive experiments prove that combining EDL loss with information maximization loss helps the model adapt better to the target domain. \\
\textbf{Acknowledgment:} Shivangi acknowledges the support of Visvesvaraya PhD fellowship by Ministry of Electronics and Information Technology (MeitY), Government of India. Rini acknowledges the financial support provided by the Department of Science and Technology, Government of India, through the WISE Post-Doctoral Fellowship Programme (Reference No. DST/WISE-PDF/ET-33/2023).



{
\bibliographystyle{plain}
\bibliography{egbib}

\begin{thebibliography}{10}

\bibitem{aguilar2023continual}
Eduardo Aguilar, Bogdan Raducanu, Petia Radeva, and Joost Van~de Weijer.
\newblock Continual evidential deep learning for out-of-distribution detection.
\newblock In {\em Proceedings of the IEEE/CVF International Conference on Computer Vision}, pages 3444--3454, 2023.

\bibitem{amini2020deep}
Alexander Amini, Wilko Schwarting, Ava Soleimany, and Daniela Rus.
\newblock Deep evidential regression.
\newblock {\em Advances in neural information processing systems}, 33:14927--14937, 2020.

\bibitem{ashfaq2023deed}
Awais Ashfaq, Markus Lingman, Murat Sensoy, and S{\l}awomir Nowaczyk.
\newblock Deed: Deep evidential doctor.
\newblock {\em Artificial Intelligence}, 325:104019, 2023.

\bibitem{bao2021evidential}
Wentao Bao, Qi~Yu, and Yu~Kong.
\newblock Evidential deep learning for open set action recognition.
\newblock In {\em Proceedings of the IEEE/CVF International Conference on Computer Vision}, pages 13349--13358, 2021.

\bibitem{caron2018deep}
Mathilde Caron, Piotr Bojanowski, Armand Joulin, and Matthijs Douze.
\newblock Deep clustering for unsupervised learning of visual features.
\newblock In {\em Proceedings of the European conference on computer vision (ECCV)}, pages 132--149, 2018.

\bibitem{chen2021source}
Cheng Chen, Quande Liu, Yueming Jin, Qi~Dou, and Pheng-Ann Heng.
\newblock Source-free domain adaptive fundus image segmentation with denoised pseudo-labeling.
\newblock In {\em Medical Image Computing and Computer Assisted Intervention--MICCAI 2021: 24th International Conference, Strasbourg, France, September 27--October 1, 2021, Proceedings, Part V 24}, pages 225--235. Springer, 2021.

\bibitem{chen2024think}
Jiayi Chen, Benteng Ma, Hengfei Cui, and Yong Xia.
\newblock Think twice before selection: Federated evidential active learning for medical image analysis with domain shifts.
\newblock In {\em Proceedings of the IEEE/CVF Conference on Computer Vision and Pattern Recognition}, pages 11439--11449, 2024.

\bibitem{deng2023uncertainty}
Danruo Deng, Guangyong Chen, Yang Yu, Furui Liu, and Pheng-Ann Heng.
\newblock Uncertainty estimation by fisher information-based evidential deep learning.
\newblock In {\em International Conference on Machine Learning}, pages 7596--7616. PMLR, 2023.

\bibitem{han2019unsupervised}
Ligong Han, Yang Zou, Ruijiang Gao, Lezi Wang, and Dimitris Metaxas.
\newblock Unsupervised domain adaptation via calibrating uncertainties.
\newblock In {\em CVPR Workshops}, volume~9, 2019.

\bibitem{he2016deep}
Kaiming He, Xiangyu Zhang, Shaoqing Ren, and Jian Sun.
\newblock Deep residual learning for image recognition.
\newblock In {\em Proceedings of the IEEE conference on computer vision and pattern recognition}, pages 770--778, 2016.

\bibitem{hoffman2018cycada}
Judy Hoffman, Eric Tzeng, Taesung Park, Jun-Yan Zhu, Phillip Isola, Kate Saenko, Alexei Efros, and Trevor Darrell.
\newblock Cycada: Cycle-consistent adversarial domain adaptation.
\newblock In {\em International conference on machine learning}, pages 1989--1998. Pmlr, 2018.

\bibitem{Kurmi_2021_WACV}
Vinod~K. Kurmi, Venkatesh~K. Subramanian, and Vinay~P. Namboodiri.
\newblock Domain impression: A source data free domain adaptation method.
\newblock In {\em Proceedings of the IEEE/CVF Winter Conference on Applications of Computer Vision (WACV)}, pages 615--625, January 2021.

\bibitem{LEE2023183}
JoonHo Lee and Gyemin Lee.
\newblock Unsupervised domain adaptation based on the predictive uncertainty of models.
\newblock {\em Neurocomputing}, 520:183--193, 2023.

\bibitem{li2024comprehensive}
Jingjing Li, Zhiqi Yu, Zhekai Du, Lei Zhu, and Heng~Tao Shen.
\newblock A comprehensive survey on source-free domain adaptation.
\newblock {\em IEEE Transactions on Pattern Analysis and Machine Intelligence}, 2024.

\bibitem{liang2020we}
Jian Liang, Dapeng Hu, and Jiashi Feng.
\newblock Do we really need to access the source data? source hypothesis transfer for unsupervised domain adaptation.
\newblock In {\em International conference on machine learning}, pages 6028--6039. PMLR, 2020.

\bibitem{litrico2023guiding}
Mattia Litrico, Alessio Del~Bue, and Pietro Morerio.
\newblock Guiding pseudo-labels with uncertainty estimation for source-free unsupervised domain adaptation.
\newblock In {\em Proceedings of the IEEE/CVF Conference on Computer Vision and Pattern Recognition}, pages 7640--7650, 2023.

\bibitem{liu2022deep}
Xiaofeng Liu, Chaehwa Yoo, Fangxu Xing, Hyejin Oh, Georges El~Fakhri, Je-Won Kang, Jonghye Woo, et~al.
\newblock Deep unsupervised domain adaptation: A review of recent advances and perspectives.
\newblock {\em APSIPA Transactions on Signal and Information Processing}, 11(1), 2022.

\bibitem{lu2023uncertainty}
Zhihe Lu, Da~Li, Yi-Zhe Song, Tao Xiang, and Timothy~M Hospedales.
\newblock Uncertainty-aware source-free domain adaptive semantic segmentation.
\newblock {\em IEEE Transactions on Image Processing}, 2023.

\bibitem{nau2022comparison}
Calvin Nau, Prashant Sankaran, and Katie McConky.
\newblock Comparison of parameter tuning strategies for team orienteering problem (top) solved with gurobi.
\newblock In {\em IIE Annual Conference. Proceedings}, pages 1--6. Institute of Industrial and Systems Engineers (IISE), 2022.

\bibitem{pandey2023learn}
Deep~Shankar Pandey and Qi~Yu.
\newblock Learn to accumulate evidence from all training samples: theory and practice.
\newblock In {\em International Conference on Machine Learning}, pages 26963--26989. PMLR, 2023.

\bibitem{park2023active}
Younghyun Park, Wonjeong Choi, Soyeong Kim, Dong-Jun Han, and Jaekyun Moon.
\newblock Active learning for object detection with evidential deep learning and hierarchical uncertainty aggregation.
\newblock In {\em The Eleventh International Conference on Learning Representations}, 2023.

\bibitem{peng2019moment}
Xingchao Peng, Qinxun Bai, Xide Xia, Zijun Huang, Kate Saenko, and Bo~Wang.
\newblock Moment matching for multi-source domain adaptation.
\newblock In {\em Proceedings of the IEEE/CVF international conference on computer vision}, pages 1406--1415, 2019.

\bibitem{raychaudhuri2023prior}
Dripta~S Raychaudhuri, Calvin-Khang Ta, Arindam Dutta, Rohit Lal, and Amit~K Roy-Chowdhury.
\newblock Prior-guided source-free domain adaptation for human pose estimation.
\newblock In {\em Proceedings of the IEEE/CVF International Conference on Computer Vision}, pages 14996--15006, 2023.

\bibitem{roy2022uncertainty}
Subhankar Roy, Martin Trapp, Andrea Pilzer, Juho Kannala, Nicu Sebe, Elisa Ricci, and Arno Solin.
\newblock Uncertainty-guided source-free domain adaptation.
\newblock In {\em European conference on computer vision}, pages 537--555. Springer, 2022.

\bibitem{saenko2010adapting}
Kate Saenko, Brian Kulis, Mario Fritz, and Trevor Darrell.
\newblock Adapting visual category models to new domains.
\newblock In {\em Computer Vision--ECCV 2010: 11th European Conference on Computer Vision, Heraklion, Crete, Greece, September 5-11, 2010, Proceedings, Part IV 11}, pages 213--226. Springer, 2010.

\bibitem{sensoy2018evidential}
Murat Sensoy, Lance Kaplan, and Melih Kandemir.
\newblock Evidential deep learning to quantify classification uncertainty.
\newblock {\em Advances in neural information processing systems}, 31, 2018.

\bibitem{sun2022prior}
Tao Sun, Cheng Lu, and Haibin Ling.
\newblock Prior knowledge guided unsupervised domain adaptation.
\newblock In {\em European conference on computer vision}, pages 639--655. Springer, 2022.

\bibitem{ulmer2021prior}
Dennis Ulmer, Christian Hardmeier, and Jes Frellsen.
\newblock Prior and posterior networks: A survey on evidential deep learning methods for uncertainty estimation.
\newblock {\em arXiv preprint arXiv:2110.03051}, 2021.

\bibitem{venkateswara2017deep}
Hemanth Venkateswara, Jose Eusebio, Shayok Chakraborty, and Sethuraman Panchanathan.
\newblock Deep hashing network for unsupervised domain adaptation.
\newblock In {\em Proceedings of the IEEE conference on computer vision and pattern recognition}, pages 5018--5027, 2017.

\bibitem{UPL-SFDA}
Jianghao Wu, Guotai Wang, Ran Gu, Tao Lu, Yinan Chen, Wentao Zhu, Tom Vercauteren, Sébastien Ourselin, and Shaoting Zhang.
\newblock Upl-sfda: Uncertainty-aware pseudo label guided source-free domain adaptation for medical image segmentation.
\newblock {\em IEEE Transactions on Medical Imaging}, 42(12):3932--3943, 2023.

\bibitem{xie2023dirichlet}
Mixue Xie, Shuang Li, Rui Zhang, and Chi~Harold Liu.
\newblock Dirichlet-based uncertainty calibration for active domain adaptation.
\newblock {\em arXiv preprint arXiv:2302.13824}, 2023.

\bibitem{xu2022deep}
Shaoxun Xu, Yufei Chen, Chao Ma, and Xiaodong Yue.
\newblock Deep evidential fusion network for medical image classification.
\newblock {\em International Journal of Approximate Reasoning}, 150:188--198, 2022.

\bibitem{yang2021transformer}
Guanglei Yang, Hao Tang, Zhun Zhong, Mingli Ding, Ling Shao, Nicu Sebe, and Elisa Ricci.
\newblock Transformer-based source-free domain adaptation.
\newblock {\em arXiv:2105.14138}, 2021.

\bibitem{zong2024dirichlet}
Chen-Chen Zong, Ye-Wen Wang, Ming-Kun Xie, and Sheng-Jun Huang.
\newblock Dirichlet-based prediction calibration for learning with noisy labels.
\newblock In {\em Proceedings of the AAAI Conference on Artificial Intelligence}, volume~38, pages 17254--17262, 2024.

\end{thebibliography}
}

\end{document}




\section{Other SFDA Technique}
\subsection{TransDA}In the paper\cite{yang2021transformer}, transformer-based network for Source-Free Domain Adaptation (SFDA) is utilized to enhance the model's ability to focus on objects and significantly improve its generalization performance.  Additionally, a novel self-supervised knowledge distillation method is explored that aids the Transformer in concentrating on target objects, further improving the model's effectiveness.

We leverage our methodology into this SFDA technique (TransDA) and named it as E-TransDA.
An evidence-based model training is implemented, assigning an input sample to the class with the
strongest evidence along with  calibrated softmax function for predicting probability and its integration into information maximization has been explored. Extensive experiments prove that combining EDL loss with information maximization loss helps the model adapt better to the target domain.
The result for the same is reported in the Table \ref{tab:result_TransDA_office31} on Office-31 dataset. We observe a consistent improvement in accuracy across all domains and in average adaptation accuracy as
well, thus validating the generalizability of our approach.

\vspace{-0.6em}
\renewcommand{\thetable}{11}
\begin{table}[ht]
    \caption{ Classification accuracies on Office31 (E-TransDA)}
    \label{tab:result_TransDA_office31}
    \centering
    \resizebox{\linewidth}{!}{%
        \begin{tabular}{c|cccccc|c}
        \toprule
        Method & A$\rightarrow$D & A$\rightarrow$W & D$\rightarrow$A & D$\rightarrow$W & W$\rightarrow$A & W$\rightarrow$D & Avg. \\
        \midrule
        TransDA \cite{yang2021transformer} & 97.20 & 95.00 & 73.70 & 99.30 & 79.30 & 99.60 & 90.68 \\
        \textbf{E-TransDA (Ours)} & \textbf{97.59} & \textbf{96.48} & \textbf{74.51} & 98.87 & \textbf{80.09} & \textbf{100.00} & \textbf{91.26} \\
        \bottomrule
        \end{tabular}
    }
\end{table}

\vspace{-1.2em}
\section{Calibration Analysis}An empirical analysis that quantifies how well the predicted confidence levels correspond to actual outcomes across the entire model, specifically, Expected Calibration Error (ECE) and Negative log likelihood (NLL) for EKS is done. The Expected Calibration Error (ECE) \cite{ye2022uncertainty},  gives the measurement of the calibration based on the difference of accuracy and confidence. 
 Confidence is the max of the calibrated target (EKS) model $f_T(x_j)$ where x_j$\in
 $\lbrace x_i^t\rbrace _{i=1}^{n_t}$

\vspace{-1em}
\begin{equation}
\text{confidence}(B_i) = \frac{1}{|B_i|} \sum_{j \in B_i} \max f_T(x_j^t)
\end{equation}

\vspace{-1.5em}
\begin{equation}
\text{accuracy}(B_i) = \frac{1}{|B_i|} \sum_{j \in B_i} 1[y_j \in \arg \max f_T(x_j^t)]
\end{equation}

\vspace{-1.8em}
\begin{equation}
\text{ECE} = \sum_{i=1}^{m} \frac{|B_i|}{n_t} \left|\text{accuracy}(B_i) - \text{confidence}(B_i)\right|
\end{equation}

\vspace{-0.6em}

where $B_i$ is the number of predictions in bin $i$, $f_T(x_j^t)$ is the calibrated target model ($ j \in [1,n_t] $), $y_j$ is the lables,
$n_t$ is the total number of data point, and $m$ is the number of bins (We take m  = 15). 

Negative Log Likelihood (NLL)\cite{yao2020negative} is a measure of how well a probability distribution predicted by a model aligns with the true distribution of the data.

 Table \ref{tab:ece_result_domainnet40} shows the results of the ECE and NLL  on Domainnet40 dataset.

\renewcommand{\thetable}{10}
\begin{table}[!h]
    \caption{ECE and NLL on Domainnet40}
    \label{tab:ece_result_domainnet40}
    \centering
    \resizebox{\linewidth}{!}{%
    \begin{tabular}{c|c|c|cccccccccccc|c}
    \toprule
        Metric & Method & $\kappa$ & C$\rightarrow$S & C$\rightarrow$P & C$\rightarrow$R & R$\rightarrow$S & R$\rightarrow$C & R$\rightarrow$P & S$\rightarrow$C & S$\rightarrow$P & S$\rightarrow$R & P$\rightarrow$S & P$\rightarrow$C & P$\rightarrow$R & Avg. \\
        \midrule
        \multirow{2}{*}{\textbf{ECE}} 
        & KSHOT\cite{sun2022prior} & BR  & 0.153 & 0.116 & 0.055 & 0.149 & 0.065 & 0.092 & 0.104 & 0.141 & 0.085 & 0.148 & 0.083 & 0.055 & 0.104 \\
        & \textbf{EKS} & BR  & \textbf{0.116} & \textbf{0.089} & \textbf{0.045} & \textbf{0.130} & \textbf{0.044} & \textbf{0.071} & \textbf{0.083} & \textbf{0.106} & \textbf{0.063} & \textbf{0.121} & \textbf{0.062} & \textbf{0.044} & \textbf{0.081} \\
        \midrule
        \multirow{2}{*}{\textbf{NLL}} 
        & KSHOT\cite{sun2022prior} & BR  & 1.338 & 1.080 & 0.495 & 1.325 & 0.741 & 0.881 & 1.036 & 1.311 & 0.753 & 1.316 & 0.935 & 0.477 & 0.974 \\
        & \textbf{EKS} & BR  & \textbf{1.194} & \textbf{0.972} & \textbf{0.460} & \textbf{1.249} & \textbf{0.734} & \textbf{0.832} & \textbf{0.989} & \textbf{1.108} & \textbf{0.594} & \textbf{1.214} & \textbf{0.892} & \textbf{0.439} & \textbf{0.890} \\
        \bottomrule
    \end{tabular}
    } 
    \vspace{-1.5em}
\end{table}

\noindent {\bf Calibration curve:} Figure \ref{fif_ECE_overall} shows the ECE curve for visualizing calibration, where the dashed line $(x = y)$ represents ideal calibration. A curve closer to this line indicates better calibration. In Figure \ref{fig:ECE_EKS_KSHOT}, the EKS curve is closer to the dashed line compared to KSHOT, while Figure \ref{fig:ECE_ES_SHOT} shows the ES curve is closer to the dashed line than SHOT, thus, validating our approach.

\vspace{-1.0em}

\renewcommand{\thefigure}{3}

\begin{figure}[!h]
  \centering

  \begin{subfigure}{0.49\linewidth}
    \includegraphics[width=\linewidth]{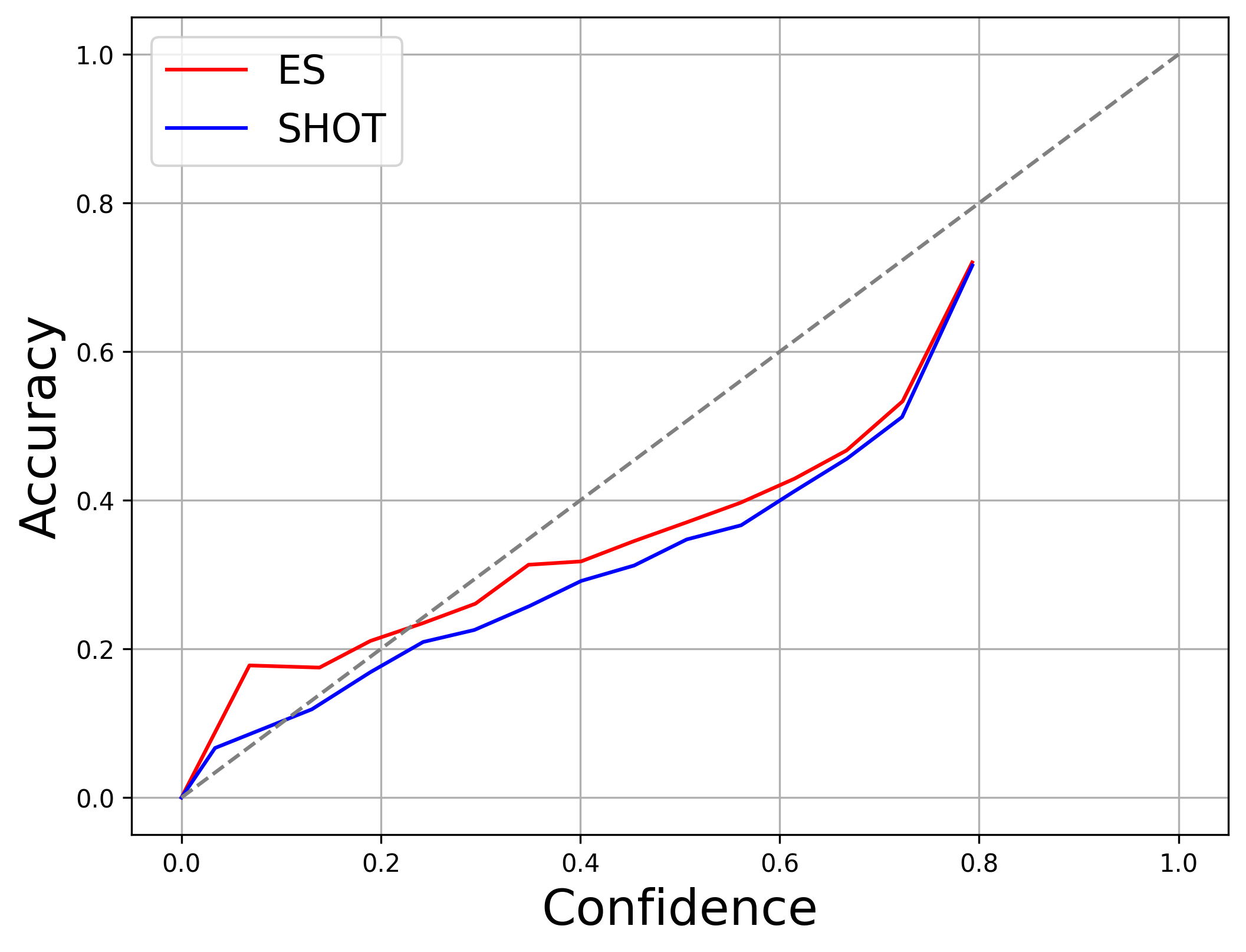}
    \caption{ES vs SHOT}
    \label{fig:ECE_ES_SHOT}
  \end{subfigure}
    \hfill
  \begin{subfigure}{0.49\linewidth}
    \includegraphics[width=\linewidth]{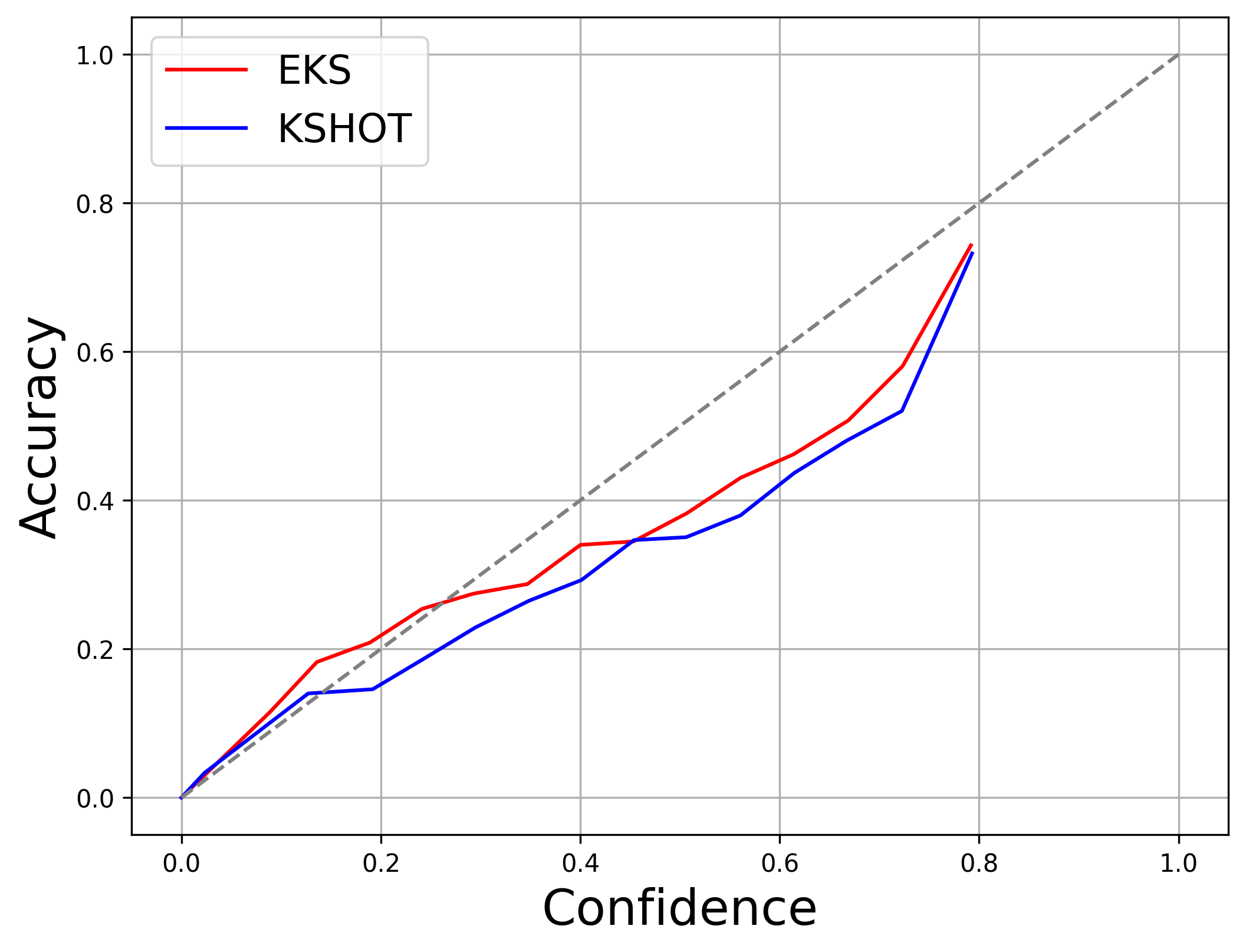}
    \caption{EKS vs KSHOT}
    \label{fig:ECE_EKS_KSHOT}
  \end{subfigure}
    \vspace{0.35em}
    \caption{Expected Calibration Error (ECE) curve  on the Domainnet40 dataset.}
    \vspace{-1em}
  \label{fif_ECE_overall}
\end{figure}

{\small
\bibliographystyle{ieee_fullname}
\bibliography{egbib}
}